\documentclass[graybox]{svmult}

\usepackage[numbers]{natbib}

\usepackage{mathptmx}       
\usepackage{helvet}         
\usepackage{courier}        
\usepackage{type1cm}        
\usepackage[bottom]{footmisc}
\usepackage{makeidx}         
\usepackage{graphicx}        
\usepackage{multicol}        
\usepackage[bottom]{footmisc}
\usepackage{amsmath}
\usepackage{amssymb}
\usepackage{amsfonts}

\usepackage{amsthm}
\usepackage{stackengine}
\usepackage{wrapfig}

\newcommand{\RR}{\mathbb{R}}

\newtheoremstyle{nospace}
{10pt}   				
{10pt}   				
{\itshape}  			
{} 		  		    	
{\bfseries} 			
{.}         			
{5pt plus 1pt minus 1pt}
{}          			

 \theoremstyle{nospace} 
 \theoremstyle{nospace} \newtheorem{discussion}{Discussion}

 \usepackage{titlesec}
 \titlespacing*{\subsection}{0pt}
     {1\baselineskip plus 0.3\baselineskip minus 0.3\baselineskip}
     {1\baselineskip plus 0.3\baselineskip minus 0.3\baselineskip}
 \titlespacing*{\section}{0pt}
     {1\baselineskip plus 0.3\baselineskip minus 0.3\baselineskip}
     {1\baselineskip plus 0.3\baselineskip minus 0.3\baselineskip}

\makeindex             

\begin{document}

\title*{How Should a Robot Assess Risk? \\ Towards an Axiomatic Theory of Risk in Robotics}
\author{Anirudha Majumdar and Marco Pavone}
\institute{Anirudha Majumdar (ani.majumdar@princeton.edu) \at Department of Mechanical and Aerospace Engineering, Princeton University, NJ 08544, USA. Work performed as a postdoctoral scholar at Stanford University.
\and Marco Pavone (pavone@stanford.edu) \at Department of Aeronautics and Astronautics, Stanford University, CA 94305, USA.}

%
%
\maketitle

\vspace{-25pt}

\abstract*{Endowing robots with the capability of assessing risk and making risk-aware decisions is widely considered a key step toward ensuring safety for  robots operating under uncertainty. But, how should a robot {\em quantify} risk? A natural and common approach  is to consider the framework whereby costs are assigned to stochastic outcomes--an assignment captured by a cost random variable. Quantifying risk then corresponds to evaluating a \emph{risk metric}, i.e., a mapping from the cost random variable to a real number. Yet, the question of what constitutes a ``good" risk metric has received  little attention within the robotics community. The goal of this paper is to explore and partially address this question by advocating \emph{axioms} that risk metrics in robotics applications should satisfy in order to be employed as rational assessments of risk. We discuss general representation theorems that precisely characterize the class of metrics that satisfy these axioms (referred to as {\em distortion risk metrics}), and provide instantiations that can be used in applications. We further discuss pitfalls of commonly used risk metrics in robotics, and discuss additional properties that one must consider in sequential decision making tasks. Our hope is that the ideas presented here will lead to a foundational framework for quantifying risk (and hence safety) in robotics applications. 
}

\abstract{Endowing robots with the capability of assessing risk and making risk-aware decisions is widely considered a key step toward ensuring safety for  robots operating under uncertainty. But, how should a robot {\em quantify} risk? A natural and common approach  is to consider the framework whereby costs are assigned to stochastic outcomes--an assignment captured by a cost random variable. Quantifying risk then corresponds to evaluating a \emph{risk metric}, i.e., a mapping from the cost random variable to a real number. Yet, the question of what constitutes a ``good" risk metric has received  little attention within the robotics community. The goal of this paper is to explore and partially address this question by advocating \emph{axioms} that risk metrics in robotics applications should satisfy in order to be employed as rational assessments of risk. We discuss general representation theorems that precisely characterize the class of metrics that satisfy these axioms (referred to as {\em distortion risk metrics}), and provide instantiations that can be used in applications. We further discuss pitfalls of commonly used risk metrics in robotics, and discuss additional properties that one must consider in sequential decision making tasks. Our hope is that the ideas presented here will lead to a foundational framework for quantifying risk (and hence safety) in robotics applications. 
}

\section{Introduction}
\label{sec:introduction} 

Safe planning and decision making under uncertainty are widely regarded as central challenges in enabling robots to successfully operate in real-world environments. By far the most common conceptual framework for addressing these challenges is to assign costs to stochastic outcomes and then to use the {\em expected value} of the resulting cost distribution as a quantity that ``summarizes" the value  of a decision.  Such a quantity can then be optimized, or bounded within a constrained formulation. However, in settings where risk has to be accounted for, this choice is rarely well justified beyond the mathematical convenience it affords. 
For example, imagine a safety-critical application such as autonomous driving; would a passenger riding in an autonomous car be happy to do so if she was told that the average behavior of the car is not to crash? While one can introduce some degree of risk sensitivity (i.e., sensitivity to the tails of the cost distribution) in the expected cost framework by simply shaping the cost function, this can quickly turn into an exercise in ``cost function hacking". Unless one is careful about the way one shapes the cost function, this can lead to the robot behaving in an irrational manner \cite{AmodeiOlahEtAl2016}. The common alternative approach, aimed at promoting risk sensitivity, is to consider a \emph{worst-case assessment} of the distribution of stochastic outcomes. In practice, however, such an assessment can often be quite conservative: an autonomous car whose goal is to never crash would never leave the garage. 

The expected value operator and the worst-case assessment are examples of \emph{risk metrics}. Informally, a risk metric is a mapping from a random variable corresponding to costs to a real number. The expected cost corresponds to risk neutrality while the worst-case assessment corresponds to extreme risk aversion. For practical applications, we would like to explore risk metrics that lie in between these extremes. This raises the following question: is there a class of risk metrics that lie between these extremes while still ensuring that the robot quantifies risk (and hence safety) in a rational and trustworthy manner? This question is central to the problem of decision making under uncertainty since the choice of a risk metric is one that must be made in \emph{any} framework that assigns costs to outcomes. Yet, despite the role of safe decision making under uncertainty as a core theme in practically all areas of robotics, this question has received very little attention within the robotics community. As a result, there is arguably no firm theoretical foundation for making an informed decision about what risk metric to use for a given robotics application. 

The goal of this paper is to provide a first step towards such a principled framework. More precisely, we describe axioms (properties) that a risk metric employed by a robot should satisfy in order to be considered sensible. To our knowledge, this is the first attempt to provide such an axiomatic framework for evaluating risk in robotics applications. Our effort is inspired by a similar effort in the finance community that led to the identification of \emph{coherent risk metrics} \cite{ArtznerDelbaenEtAl1999, ShapiroDentchevaEtAl2014} as a class of risk metrics that have desirable properties for assessing the risk associated with a financial asset (e.g., a portfolio of stocks). The influence that these ideas have had can be gauged by the fact that in 2014 the Basel Committee on Banking Supervision changed its guidelines for banks to replace the Value at Risk (VaR) (a \emph{non-}coherent risk metric) with the Conditional Value at Risk (CVaR) (a coherent risk metric) for assessing market risk \cite{BankingSupervision2014}.

We believe that the question of what properties a risk metric should satisfy in robotics applications is a fundamental one: a robot's inability to assess risks in a rational way could lead to behavior that is harmful both to itself and humans or other autonomous agents around it. Our hope is that the ideas presented in this paper can help the community converge upon a set of properties that risk metrics must satisfy in order for the robot's decision-making system to be considered rational and trustworthy, paralleling a similar effort in the financial industry. Indeed, it is conceivable that in the not-so-distant future, robots such as autonomous cars or unmanned aerial vehicles (UAVs) deployed in safety-critical scenarios will be subject to regulatory frameworks that mandate the use of ``officially-approved" risk metrics. 


{\bf Related Work:} Our work here is inspired by the efforts over the last two decades towards the development of an axiomatic theory of risk in finance and operations research. These efforts have resulted in widespread acceptance of the class of coherent risk metrics as ``rational" measures of risk for financial assets. Coherent risk metrics are defined by four axioms that any sensible assessment of financial risk must satisfy (see Section \ref{sec:DRMs}). Additional axioms beyond the four characterizing coherent risk metrics have also been studied and lead to more refined classes of risk metrics. We refer the reader to \cite{FollmerSchied2011, ShapiroDentchevaEtAl2014} for an introduction to this vast literature. Our main contribution here is to parallel the efforts in the finance community on proposing and characterizing axioms that any sensible assessment of risk in a robotics application should satisfy. We propose these axioms in Section \ref{sec:axioms section} and provide interpretations for them in a robotics context. As we will see, the class of risk metrics fulfilling these axioms corresponds precisely to the class of \emph{distortion risk metrics}, which form a subset of coherent risk metrics and have previously been studied in the finance literature \cite{Wang2000,IancuPetrikEtAl2015,FollmerSchied2011, BertsimasBrown2009b}.

As discussed previously, the most commonly used risk metrics in robotics are the expected cost and worst-case metrics. 
While these risk metrics are justifiable in certain contexts (e.g., expected cost in applications where the distribution of costs is known to not have a long tail or worst-case assessments in low-level control tasks such as trajectory tracking), many robotics applications call for a more nuanced assessment of risk. \emph{Chance constrained programming} \cite{CharnesCooper1959} provides one avenue towards such assessments. In particular, a chance constraint specifies an upper bound on the probability of incurring a cost higher than a given threshold. Chance constraints have been widely studied in robotics for motion planning under uncertainty \cite{BlackmoreOnoEtAl2011, DuToitBurdick2011, OnoPavoneEtAl2015}. While chance constraints are suitable for capturing risk corresponding to \emph{boolean} events (e.g., collisions with obstacles), they do not take into account variations in the tails of cost distributions (since they are not affected by changes to the value of the cost above the given threshold). Chance constraints are thus not suitable for capturing risk in settings where one must consider a range of cost outcomes (in contrast to boolean events). 

Another popular way to quantify risk in control theory and decision making is through the notion of \emph{distributional robustness} \cite{DelageYe2010, XuMannor2010, SummersWarringtonEtAl2015}. Distributional robustness captures the idea that the underlying distribution from which random outcomes of the world are generated may itself be uncertain. Such scenarios are referred to as \emph{ambiguous} in the literature on human decision making \cite{GilboaMarinacci2016}. In such scenarios, while the precise distribution may be unknown, one may know certain properties of the underlying distribution (e.g., its first few moments, or that it lies in a given set of possible distributions). One can then compute the worst-case expectation of the cost function over the \emph{set of distributions} that satisfy the known properties (e.g., the set of all distributions that have the given moments). As we will see in Section \ref{sec:DRMs}, the set of risk metrics fulfilling the axioms advocated in this paper have an interpretation in terms of distributional robustness. However, not all distributionally robust risk metrics satisfy the proposed axioms.

Beyond robotics and finance, the notion of risk is of central concern to the theory of human decision making under uncertainty. A historically prominent theory is \emph{subjective expected utility (SEU)} theory \cite{NeumannMorgenstern1944}, where humans make decisions that maximize the expected value of a utility function. This is analogous to the method for taking into account risk aversion in autonomous agents by ``shaping" the cost function. However, SEU theory is inconsistent with a number of experimental observations \cite{Ellsberg1961, KahnemanTversky1979, Allais1990}. In particular, SEU theory does not take into account the experimentally observed fact that humans are \emph{ambiguity averse} \cite{Ellsberg1961, Allais1990}, i.e., averse to situations where there is uncertainty about the underlying probability distribution from which outcomes are drawn (previously discussed in the context of distributional robustness). Models of ambiguity aversion include \emph{prospect theory}, which has risen to prominence over the last several decades, along with more recent and sophisticated theories \cite{GilboaMarinacci2016} that model the human as minimizing a distributionally robust risk metric.
These theories are thus closely related to the class of distortion risk metrics (which we advocate for in this paper for use in robotics) since they can also be interpreted as a special class of distributionally robust risk metrics. 


{\bf Outline:} The outline of this paper is as follows. Section \ref{sec:risk section} formally introduces the notion of a risk metric (Section \ref{sec:risk metrics}) and proposes an interpretation of risk in robotics applications (Section \ref{sec:interpretation}). In Section \ref{sec:axioms section} we advocate axioms that risk metrics in robotics applications should satisfy in order to be considered sensible (Sections \ref{sec:axioms}), provide examples of metrics that satisfy them, and discuss pitfalls stemming from  using risk metrics that do not satisfy these axioms (Section \ref{sec:examples}). Section \ref{sec:DRMs} discusses representation theorems that precisely characterize the class of risk metrics fulfilling the axioms we propose. Section \ref{sec:multiperiod} proposes additional properties that one must consider in sequential decision making tasks. Section \ref{sec:discussion} concludes the paper with directions for future research. We note that we highlight a number of {\em points of discussion} throughout the paper. Our hope is that these will form the basis for discussion and debate at the Blue Sky session during the conference and will provoke new directions for future research.

\section{Assessing Risk: Preliminaries}
\label{sec:risk section}

In this section, we formally introduce risk metrics and propose an intuitive interpretation of risk quantification in robotics contexts. This interpretation will form the basis for the axioms we advocate in Section \ref{sec:axioms section}.

\subsection{Risk Metrics} 
\label{sec:risk metrics}

We denote the set of possible outcomes that may occur when a robot operates in uncertain settings as $\Omega$. In order to avoid heavy use of measure theoretic notions, we take $\Omega$ to be finite. Denote by $\mathbb{P}$ a probability mass function that assigns probabilities $\mathbb{P}(\omega)$ to outcomes $\omega \in \Omega$. Consider a cost function $Z: \Omega \rightarrow  \mathbb{R}$ that assigns costs $Z(\omega)$ to outcomes. The cost $Z$ is then a random variable, namely the cost random variable. Let $\mathcal{Z}$ denote the set of all random variables on $\Omega$. A \emph{risk metric} is a mapping $\rho: \mathcal{Z} \rightarrow \RR$. In other words, a risk metric maps a cost random variable to a real number. 


\subsection{Interpretation of Risk in Robotics Applications}
\label{sec:interpretation}

Imagine a fictional government agency known as the Robot Certification Agency (RCA) that is responsible for certifying if a given robot is safe to operate in the real world. How should the RCA quantify the risk faced by this robot? As an example, consider an autonomous car driving from one city to another. While performing this task, the robot will incur random costs $Z$ (e.g., due to fuel consumption, time, crashes, mechanical wear and tear, etc.). In order to provide clear interpretations of the axioms for risk metrics discussed below, we will take the following axiom as a key starting point.

\vspace{5pt}

\noindent {\bf A0. Monetary costs.} The costs $Z$ are expressed in monetary terms.


\vspace{5pt}

This axiom ensures that the costs assigned to outcomes have a tangible and interpretable value, which will be instrumental in defining a meaningful notion of risk below. Such an axiom may also provide a handle on reasoning about insurance policies for safety-critical robots (e.g., autonomous cars).
We note that our starting point contrasts with one where one considers a more abstract and subjective notion of cost (e.g., quadratic state and control costs for Linear Quadratic Regulator problems).

Given this assumption,  suppose that the RCA demands that the robot's owner must deposit an amount of money $\rho(Z)$ before the robot is deployed such that the RCA is satisfied that the owner will be able to cover the potential costs incurred during operations (e.g., making repairs to the robot due to an accident) with the amount $\rho$. We define the amount $\rho(Z)$ as the perceived \emph{risk} from operating the robot. The particular risk metric $\rho: \mathcal{Z} \rightarrow \RR$ the RCA uses will depend on its attitude towards risk and may depend on the application under consideration. For example, the RCA may ask for a deposit $\rho(Z) = \mathbb{E}[Z]$ if it is risk neutral. If it wants to be highly conservative, the RCA may demand a deposit equal to the worst-case cost outcome. The question we will pursue in Section \ref{sec:axioms section} is the following: what properties must $\rho$ satisfy in order for it to be considered sensible?

We note that we are using the RCA here as a pedagogical tool to provide an interpretation of risk in robotics applications and to motivate the axioms described in Section \ref{sec:axioms section}. In reality, the robot's decision-making system will assess risks and make decisions based on those assessments. 

\section{An Axiomatization of Risk Metrics for Robotics Applications}
\label{sec:axioms section}

\subsection{Axioms and their Interpretations}
\label{sec:axioms}

We now parallel results in finance \cite[Chapter 4]{FollmerSchied2011} \cite{ShapiroDentchevaEtAl2014} and propose six axioms for risk metrics. Specifically, we make the case that these axioms should be fulfilled by any risk metric used in a robotics application in order for it to be considered a sensible assessment of risk. For each axiom, we first  provide a formal statement and then an intuitive interpretation based on the interpretation of risk from Section \ref{sec:interpretation}.

\quad

\begin{figure*}[t!]
  \begin{center}
    \includegraphics[trim = 0mm 0mm 0mm 0mm, clip, width=0.5\textwidth]{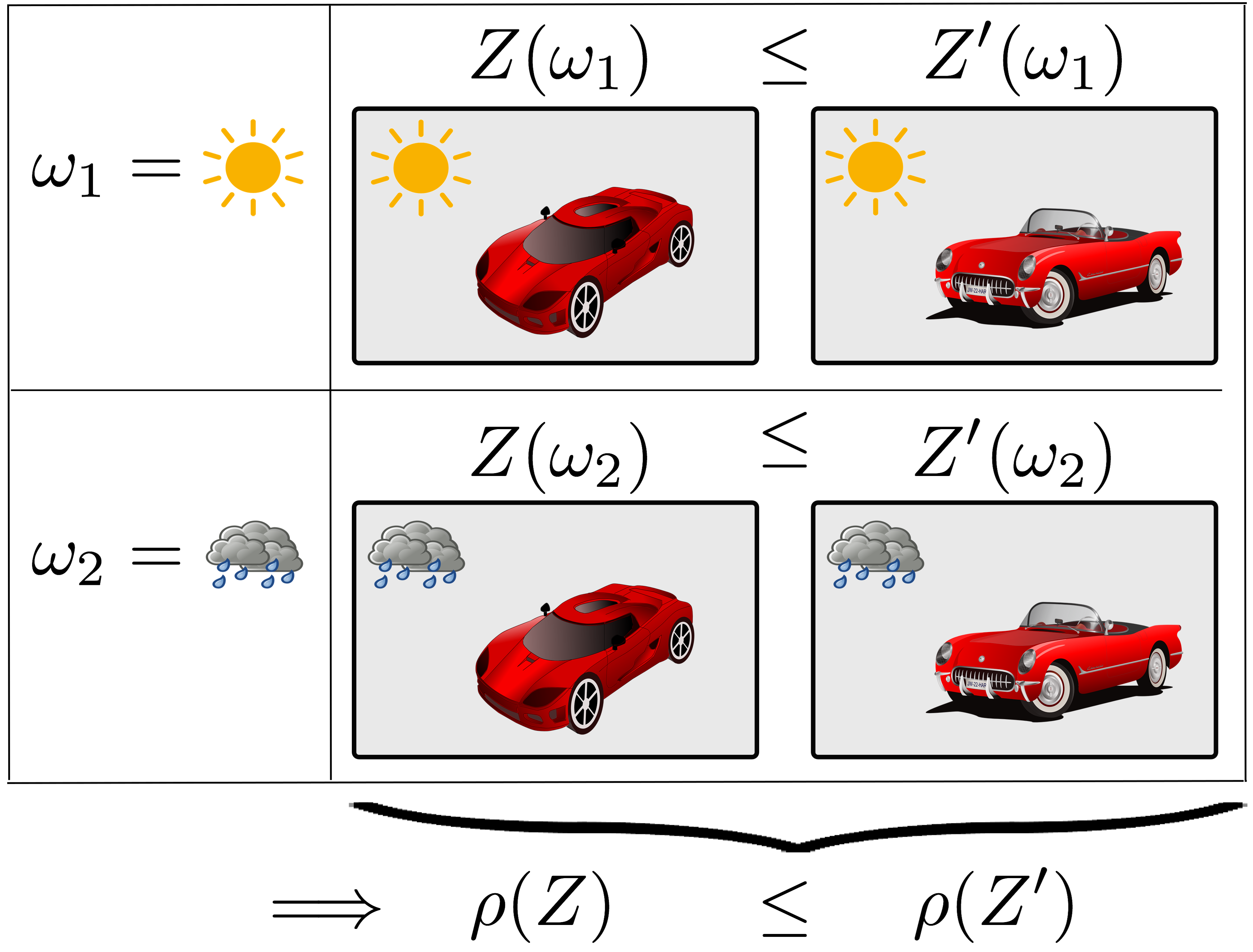}
  \end{center}
  \caption{Monotonicity. Consider two robot cars that incur random costs $Z$ and $Z'$ respectively. The situation consists of two outcomes $\omega_1$ and $\omega_2$ corresponding to sunny and rainy weather, respectively. If the car that incurs costs $Z$ (first column) incurs a lower cost than the other car (second column) no matter what the weather is, monotonicity states that it should be considered less risky.
  \label{fig:monotonicity}}
\end{figure*}

\noindent {\bf A1. Monotonicity.} Let $Z, Z' \in \mathcal{Z}$ be two cost random variables. Suppose $Z(\omega) \leq Z'(\omega)$ for all $\omega \in \Omega$. Then $\rho(Z) \leq \rho(Z')$. 

\noindent {\bf Interpretation:} If a random cost $Z'$ is guaranteed to be greater than or equal to a random cost $Z$ \emph{no matter what random outcome occurs}, then $Z'$ must be deemed at least as risky as $Z$. One can think of the random costs as corresponding to two different robots, or the same robot performing different tasks, or executing different controllers. Given our interpretation of risk, this axiom states that the RCA must demand at least as large a deposit for covering costs for the robot (or task) corresponding to $Z'$ as for the robot (or task) corresponding to $Z$. This is a sensible requirement since we are guaranteed to incur at least as high a cost in the second scenario as the first no matter which outcome $\omega \in \Omega$ is realized. An example is illustrated in Figure \ref{fig:monotonicity}, where we have two robot cars corresponding to $Z$ and $Z'$ and two outcomes $\omega_1$ and $\omega_2$ corresponding to sunny and rainy weather, respectively. If the car corresponding to $Z$ incurs a lower cost no matter what the weather is, monotonicity states that it should be considered less risky.

\noindent {\bf A2. Translation invariance.} Let $Z \in \mathcal{Z}$ and $c \in \RR$. Then $\rho(Z+c) = \rho(Z) + c$. 

 \noindent {\bf Interpretation:} If one is charged a deterministic cost $c$ (in addition to the random costs incurred when the robot is operated), then the RCA should demand that this amount $c$ be set aside in addition to money for covering the other costs from operating the robot:
   \begin{figure*}[h!]
  \begin{center}
    \includegraphics[trim = 0mm 0mm 0mm 0mm, clip, width=0.5\textwidth]{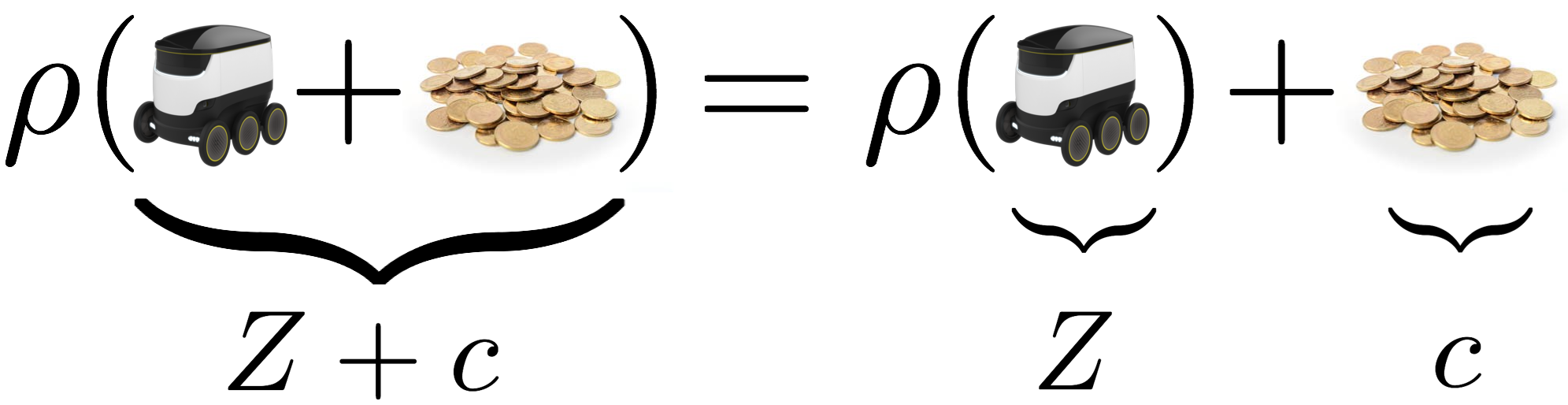}
  \end{center}
  \label{fig:translation invariance}
\end{figure*}

\noindent Note that this axiom also implies that $\rho(Z - \rho(Z)) = 0$. Thus, $\rho(Z)$ is the smallest amount that must be deducted from the costs in order to make the task risk-free.
 
\noindent {\bf A3. Positive homogeneity.} Let $Z \in \mathcal{Z}$ and $\beta \geq 0$ be a scalar. Then $\rho(\beta Z) = \beta \rho(Z)$. 
 
\noindent {\bf Interpretation:} If all the costs incurred by the robot (regardless of the random outcome) are scaled by $\beta$, the RCA demands that the deposit is scaled commensurately. This is reasonable since this corresponds to simply changing the units of cost (recall that we assumed that the costs are expressed in monetary terms).


 \noindent {\bf A4. Subadditivity.} Let $Z, Z' \in \mathcal{Z}$. Then $\rho(Z + Z') \leq \rho(Z) + \rho(Z')$.
 
 \noindent {\bf Interpretation:} This axiom encourages diversification of risk. For example, imagine a system with two robots. Suppose that $Z$ and $Z'$ are costs incurred by Robot 1 and Robot 2 respectively. The left-hand side (LHS) of the inequality corresponds to the deposit that the RCA demands when both robots are run simultaneously, while the right-hand side  (RHS) corresponds to the sum of the deposits when the robots are deployed separately. Axiom A4 then states that deploying both robots simultaneously is at most as risky as deploying them separately:
 
  \begin{figure*}[h!]
  \begin{center}
    \includegraphics[trim = 0mm 0mm 0mm 0mm, clip, width=0.5\textwidth]{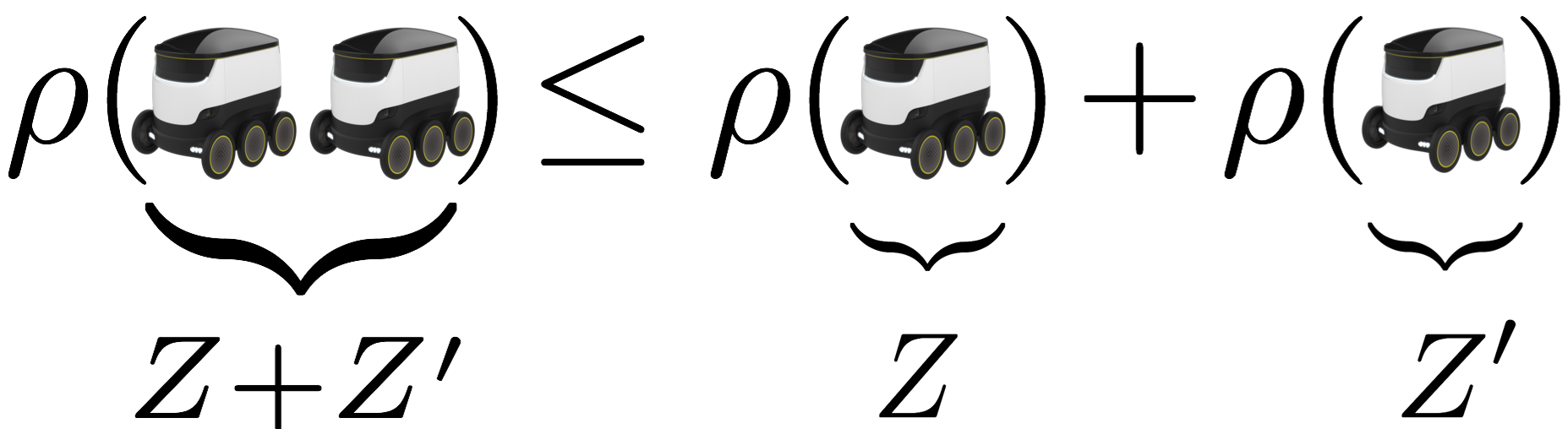}
  \end{center}
  \label{fig:subadditivity}
\end{figure*}
 This captures the intuition that one robot acts as a hedge against the other robot failing (i.e., if one of the robot fails in some way, then the other will make up for this loss). Another interpretation is that this axiom promotes redundancy in the system. The exact interpretation of the LHS and RHS of the inequality in A4 will depend on the particular application under consideration. For example, the two sub-costs $Z$ and $Z'$ may correspond to two separate sub-tasks that the robot must perform. In this case,  the LHS corresponds to the robot performing both tasks simultaneously while the RHS corresponds to performing them independently. A4 encodes the intuition that performing both tasks simultaneously is less risky since one sub-task can act as a hedge against the other. 

We note that A3 and A4 together imply convexity: 
$$\rho(\lambda Z + (1 - \lambda) Z') \leq \lambda \rho(Z) + (1 - \lambda) \rho(Z'), \text{ for all } \lambda \in [0,1].$$
  
 \noindent {\bf A5. Comonotone additivity.} Suppose $Z$ and $Z'$ are comonotone, i.e., $(Z(\omega) - Z(\omega'))(Z'(\omega) - Z'(\omega')) \geq 0$, for all $(\omega,\omega') \in \Omega \times \Omega$. Then $\rho(Z + Z') = \rho(Z) + \rho(Z')$. 
 
\noindent {\bf Interpretation:} This axiom supplements A4. In particular, A5 states that if two costs rise and fall together, then there is no benefit from diversifying (e.g., if one robot always performs poorly at a task when the other does or when a robot performs poorly at a sub-task when it also performs poorly at another one). 
  

\noindent {\bf A6. Law invariance.} Suppose that $Z$ and $Z'$ are identically distributed. Then $\rho(Z) = \rho(Z')$.
 
 \noindent {\bf Interpretation:} If two tasks have the same distribution of costs, then the RCA demands an equal deposit in both cases. For example, suppose $\Omega = \{\omega, \omega' \}$ with both outcomes having probability 0.5. Further, suppose $Z(\omega) = 1$, $Z(\omega') = 10$, $Z'(\omega) = 10$, $Z'(\omega') = 1$. The two situations must be considered equally risky even though the assignment of costs to events is different. 
 
\vspace{5pt} 
 
Taken together, Axioms A1 -- A6 capture a fairly exhaustive set of essential properties that we believe any reasonable quantification of risk in robotics should obey given the interpretation of risk we proposed in Section \ref{sec:interpretation}. A hypothetical RCA that quantifies risk in a manner that is consistent with these axioms would be considered a sensible one. Moreover, robots that assess risks according to risk metrics that fail to satisfy some of these axioms can behave in a manner that would be considered extremely unreasonable and arguably very unsafe, as illustrated in Section \ref{sec:examples}. We thus advocate risk metrics that satisfy Axioms A1 -- A6 for use in robotics applications.
 
\subsection{Examples and Pitfalls of Commonly Used Risk Metrics}
\label{sec:examples}

In this section, we first discuss examples of existing risk metrics that fulfill Axioms A1 -- A6. We then discuss commonly used risk metrics that do not fulfill some of these axioms, along with pitfalls stemming from their use. Collectively, the discussion provided here motivates the formal introduction in the next section of distortion risk metrics (equivalently, risk metrics satisfying A1 -- A6) as a general class of risk metrics for robotic applications.

An important risk metric that satisfies Axioms A1 -- A6 is the Conditional Value at Risk (CVaR) \cite{RockafellarUryasev2000}. The $\textrm{CVaR}_\alpha$ for a random cost $Z$ at level $\alpha$ is defined as:

\begin{equation}
\text{CVaR}_{\alpha} (Z) := \frac{1}{\alpha} \int_{1-\alpha}^{1} \textrm{VaR}_{1-\tau} (Z) \ d \tau,
\end{equation}
where $\textrm{VaR}_\alpha (Z)$ is the \emph{Value at Risk (VaR)} at level $\alpha$, i.e., simply the $(1-\alpha)$-quantile of the cost random variable  $Z$:
\begin{equation}
\label{eq:VaR}
\textrm{VaR}_\alpha (Z) := \min \{z \ | \ \mathbb{P}[Z > z] \leq \alpha \}.
\end{equation}

\begin{figure*}[h!]
  \begin{center}
    \includegraphics[trim = 0mm 0mm 0mm 0mm, clip, width=0.7\textwidth]{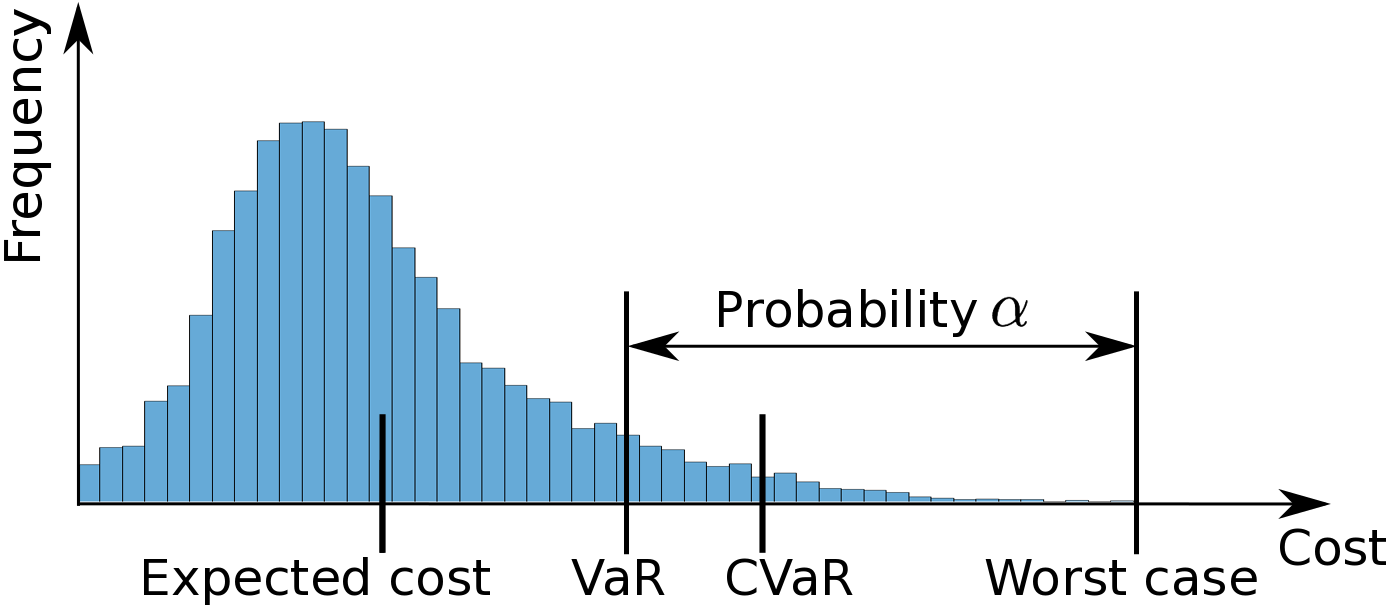}
  \end{center}
  \caption{An illustration of four important risk metrics, namely: expected cost, worst case, Value at Risk (VaR), and Conditional Value at Risk (CVaR). Intuitively, VaR is the $(1-\alpha)$-quantile of the cost distribution. CVaR is the expected value of costs in the conditional distribution of the cost distribution's upper $(1-\alpha)$-tail and is thus a metric of ``how bad is bad." The CVaR, expected cost, and worst case metrics satisfy Axioms A1 -- A6, but VaR does not. 
  \label{fig:cvar}}
\end{figure*}

\noindent Intuitively, $\text{CVaR}_{\alpha}$ is the expected value of $Z$ in the conditional distribution of $Z$'s upper $(1-\alpha)$-tail. It can thus be interpreted as a risk metric that quantifies ``how bad is bad." We note that the expected cost and worst-case assessment also satisfy A1 -- A6. Figure \ref{fig:cvar} provides a visualization of the expected cost, worst case, VaR, and CVaR. Axioms A1 -- A6 thus define a broad class of risk metrics that capture a wide spectrum of risk assessments from risk-neutral to worst-case. In Section \ref{sec:DRMs}, we will discuss theorems that allow us to precisely characterize \emph{all} risk metrics that satisfy A1 -- A6 and easily generate new examples of such metrics.

However, there are many examples of popular risk metrics that \emph{do not} fulfill the axioms we advocate. For example, a very popular metric to quantify risk in robotics applications is the mean-variance risk metric: $\mathbb{E}[Z] + \beta \textrm{Variance}[Z]$ (see, e.g., \cite{KuindersmaGrupenEtAl2013,GarciaFernandez2015,MannorTsitsiklis2011}). The mean-variance metric satisfies A6 but fails to satisfy the other axioms. This can lead to a robot that utilizes the mean-variance metric making decisions that would be considered unreasonable. Consider the setup in Table \ref{tab:mean-variance} (based on \cite{MaccheroniMarinacciEtAl2009}), where $\omega_1,\omega_2,\omega_3,\omega_4$ are disturbance outcomes, and $Z$ and $Z'$ are the costs resulting from executing two different controllers $\pi$ and $\pi'$. Which controller should the robot execute? Controller $\pi$ results in lower costs \emph{no matter what the disturbance outcome is} and hence should be preferred by any sensible decision maker. However, computing the mean-variance risk metric with $\beta = 1$, we see that:
$$\mathbb{E}[Z] +  \textrm{Variance}[Z] = 3.75 > \mathbb{E}[Z'] +  \textrm{Variance}[Z'] = 3.4375.$$
The robot would hence strictly prefer $\pi'$. This unreasonable behavior is a result of the mean-variance risk metric failing to satisfy Axiom A1 (monotonicity). 

\begin{table}
\parbox{.47\linewidth}{
\centering
\begin{tabular}{ | c | c | c | c | c |}
    \hline 
    Outcome  & $\omega_1$ & $\omega_2$ & $\omega_3$ & $\omega_4$  \\ \hline
    Probability  & 0.25 & 0.25 & 0.25 & 0.25  \\ \hline
    $Z$  & 1 & 2 & 3 & 4   \\ \hline
    $Z'$  & 2 & 2 & 3 & 4   \\ \hline
  \end{tabular}
 \caption{Issues with the mean-variance risk metric. Any rational agent would choose controller $\pi$ (with associated costs $Z$) since it results in lower costs no matter what the disturbance outcome is. But, using the mean-variance risk metric results in choosing $\pi'$ (with associated costs $Z'$).}
\label{tab:mean-variance} 
}
\hfill
\parbox{.47\linewidth}{
\centering
\vspace{-28pt}
\begin{tabular}{ | c | c | c | c |}
    \hline 
    Outcome  & $\omega_1$ & $\omega_2$ & $\omega_3$  \\ \hline
    Probability  & 0.4 & 0.4 & 0.2 \\ \hline
    $Z$  & 1 & 2 & 3  \\ \hline
    $Z'$  & 1 & 1.99 & $10^{10}$   \\ \hline
  \end{tabular}
 \caption{Issues with the Value at Risk (VaR) metric. Any reasonable agent would prefer costs $Z$ to $Z'$. However, utilizing VaR results in choosing $Z'$.}
 }
\label{tab:VaR} 
\end{table}


The Value at Risk (VaR) (defined in Equation \eqref{eq:VaR}) is another example of a risk metric that does not satisfy all the axioms (it is easily verified that VaR does not satisfy A4 (subadditivity), but satisfies the other axioms). VaR is closely related to chance constraints (see Section \ref{sec:introduction}) since the constraint $\textrm{VaR}_\alpha (Z) \leq 0$ corresponds to the chance constraint $\mathbb{P}[Z > 0] \leq \alpha$. Using the VaR metric can also lead to behavior that is arguably very unreasonable and unsafe. For example, consider the costs $Z$ and $Z'$ in Table \ref{tab:VaR}. Any reasonable agent would prefer costs $Z$ to $Z'$ due to the extremely large costs associated with $\omega_3$. However, we see that $\textrm{VaR}_{0.3} (Z) = 2$, while $\textrm{VaR}_{0.3} (Z') = 1.99$. Thus, utilizing VaR results in a strict preference for $Z'$. We note that using CVaR instead results in a preference for $Z$.


\begin{table}
\centering
 \begin{tabular}{ | c | c |}
    \hline 
      & Axioms  \\ \hline
    Conditional Value at Risk (CVaR) & A1 -- A6  \\ \hline
    Expected Cost & A1 -- A6  \\ \hline
    Worst case & A1 -- A6  \\ \hline
    Mean--Variance & A6   \\ \hline
    Entropic risk & A1, A2, A6  \\ \hline
    Value at Risk (VaR) &  A1 -- A3, A5, A6 \\ \hline
    Standard semi-deviation & A1 -- A4,  A6   \\ \hline
  \end{tabular}
\caption{Axioms satisfied by popular risk metrics.}
\label{tab:metrics} 
\end{table}

Table \ref{tab:metrics} lists the axioms satisfied by popular risk metrics in the literature (we refer the reader to \cite{ShapiroDentchevaEtAl2014} for definitions). We note that the standard semi-deviation is widely used in finance \cite{ShapiroDentchevaEtAl2014}, while the entropic risk metric has been popular in control theory for risk-averse control \cite{Whittle1981, GloverDoyle1987}.
 
\section{Distortion Risk Metrics}
\label{sec:DRMs}

\begin{figure}
\centering
  \begin{minipage}[c]{0.3\textwidth}
\caption{Any coherent risk metric (and thus any distortion risk metric) can be represented as an expectation with respect to a worst-case probability mass function, chosen adversarially from a compact convex subset (referred to as a \emph{risk envelope}) of the probability simplex.
  \label{fig:risk envelope}}
  \end{minipage}  \hspace{10pt} 
  \begin{minipage}[c]{0.3\textwidth}
    \includegraphics[trim = 0mm 0mm 0mm 0mm, clip, width=0.9\textwidth]{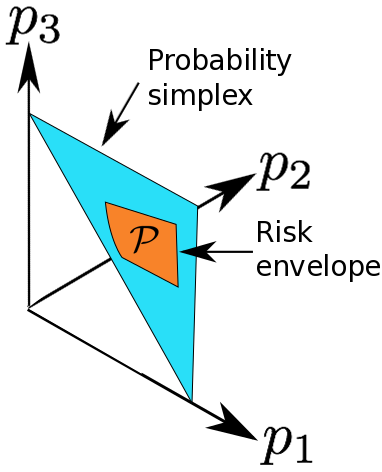}
  \end{minipage}
\end{figure}

Risk metrics satisfying Axioms A1 -- A6 have been studied in the context of portfolio optimization in finance and are known as \emph{distortion risk metrics} \cite{Wang2000,IancuPetrikEtAl2015,FollmerSchied2011, BertsimasBrown2009b}. These risk metrics are also equivalent to the class of \emph{spectral risk measures} \cite{Acerbi2002}. They enjoy an elegant characterization, which we discuss below, in terms of the CVaR metric. Before doing so, we first discuss representations for risk metrics satisfying subsets of the above axioms. In particular, Axioms A1 -- A4 correspond to those of \emph{coherent risk metrics (CRMs)}. CRMs enjoy a universal representation theorem:
\begin{equation}
\rho(Z) = \max_{p \in \mathcal{P}} \mathbb{E}_p [Z],
\end{equation}
where $\mathcal{P}$ is a compact convex set of probability mass functions. In other words, any coherent risk metric can be represented as an expectation with respect to a worst case probability mass function, chosen adversarially from a given compact convex set (referred to as a  \emph{risk envelope}). An example of a risk envelope is visualized in Figure \ref{fig:risk envelope}. Coherent risk metrics thus capture, as a by-product, the notion of \emph{distributional robustness} (see Section \ref{sec:introduction}), i.e., robustness to uncertainty over the underlying distribution itself. 

Risk metrics satisfying Axioms A1 -- A5 are known as \emph{comonotonic risk metrics} \cite[Chapter 4.7]{FollmerSchied2011}. These risk metrics have a characterization in terms of Choquet integrals, which we now discuss with the help of additional terminology.

\begin{definition}
A set function $g: 2^\Omega \rightarrow [0,1]$ is called \emph{monotone} if $g(A) \leq g(B)$ for all $A \subseteq B \subseteq \Omega$, and \emph{normalized} if $g(\varnothing) = 0$ and $g(\Omega) = 1$. If, in addition, $g$ satisfies 
$$g(A \cup B) + g(A \cap B) \leq g(A) + g(B),$$
$g$ is called \emph{submodular}. 
\end{definition}
\noindent Recall that $\Omega$ corresponds to the set of outcomes that may occur when the robot operates (see Section \ref{sec:risk metrics}). The set $2^\Omega$ is the power set (i.e., set of all subsets) of $\Omega$. We now define the Choquet Integral \cite{Choquet1954} and state a representation theorem for comonotonic risk metrics in terms of these integrals.

\begin{definition}
The \emph{Choquet Integral} of a random variable $Z \in \mathcal{Z}$ with respect to a monotone, normalized set function $g: 2^\Omega \rightarrow [0,1]$ is 

$$\int Z \ d g := \int_{-\infty}^0 (g(Z > z) - 1) \ d z + \int_0^\infty g(Z > z) \ d z.$$

\end{definition}
Here, the integrals in the RHS are Riemann integrals. Informally, the Choquet integral is a generalization of the Lebesgue integral that allows the integration operation to have a \emph{nonlinear} dependence on $Z$ (in contrast to the Lebesgue integral, which is a linear operator).
\begin{theorem}[Representation of comonotonic risk metrics \cite{Schmeidler1986}]
A coherent risk metric $\rho: \mathcal{Z} \rightarrow \mathbb{R}$ is comonotonic if and only if it can be written as a Choquet Integral $\int Z \ d g$, where $g: 2^\Omega \rightarrow [0,1]$ is a monotone, normalized, and submodular set function.
\end{theorem}


Risk metrics satisfying Axioms A1 -- A6 are known as distortion risk metrics. These metrics inherit the representation theorems for coherent and comonotonic risk metrics and enjoy a further elegant characterization in terms of the CVaR metric.

\begin{theorem}[Representation of distortion risk metrics \cite{FollmerSchied2011}]
A risk metric $\rho: \mathcal{Z} \rightarrow \mathbb{R}$ is a distortion risk metric if and only if there exists a function $\nu: [0,1] \rightarrow [0,1]$, satisfying $\int_{\alpha=0}^1 \nu(d\alpha) = 1$ (i.e., $\nu$ defines a probability measure on the set $[0,1]$), such that:
\begin{equation}
\label{eq:DRMs}
\rho(Z) = \int_{\alpha=0}^1 \text{CVaR}_{\alpha} (Z) \  \nu(d \alpha).
\end{equation}
\end{theorem}

This theorem provides us with a precise mathematical characterization of \emph{all} distortion risk metrics and allows us to generate examples of such metrics by choosing functions $\nu$ satisfying the assumptions of the theorem. 

To summarize our discussion so far, based on our arguments in Section \ref{sec:axioms} we advocate the use of distortion risk metrics (i.e., risk metrics satisfying A1 -- A6, or equivalently risk metrics of the form \eqref{eq:DRMs}) for evaluating risk in robotics applications. This is in contrast to popular risk metrics used in the robotics literature (e.g., mean-variance, or VaR) and other popular classes of risk metrics used in finance (e.g., coherent risk metrics, or comonotonic risk metrics). However, we end this section by noting the possibility that for certain applications distortion risk metrics may constitute too restrictive a class of risk metrics. We leave the following as a point for discussion and future work. 

\begin{discussion}[{\bf Axioms A1 -- A6}]
\emph{The axioms of monotonicity (A1), translation invariance (A2), positive homogeneity (A3), and law invariance (A6) should arguably be applicable in any robotics application. Subadditivity (A4) and comonotone additivity (A5) are also intuitively appealing, particularly for applications that involve some degree of high-level decision making (since a high-level decision making system should ideally diversify risks). However, the interpretation of diversification is somewhat unclear in certain applications. For example, imagine a humanoid robot whose goal is to minimize a cost function that is a sum of two components $Z$ and $Z'$, where $Z$ penalizes one aspect of the robot's motion (e.g., deviations of the robot's torso from the vertical orientation) while $Z'$ penalizes another aspect (e.g., deviation of the robot's gaze from a target). For such low-level control tasks, the interpretation of $\rho(Z) + \rho(Z')$ is not entirely clear since it is not possible to perform the different subtasks corresponding to $Z$ and $Z'$ independently of each other. The following is thus a question for discussion and future work: should A4 and A5 be abandoned (or replaced by other axioms) for such tasks?}
\end{discussion}

\section{Sequential Decision Making and Time Consistency} 
\label{sec:multiperiod}

The sequential nature of many decision-making tasks in robotics gives rise to additional important considerations beyond the ones discussed in Section \ref{sec:axioms section}. Again paralleling results in finance \cite{Shapiro2009, Ruszczynski2010, ShapiroDentchevaEtAl2014}, we discuss properties that ensure the temporal consistency of risk assessments in sequential decision-making tasks.


\begin{wrapfigure}{R}{0.35\textwidth}
\centering
\includegraphics[width=0.35\textwidth]{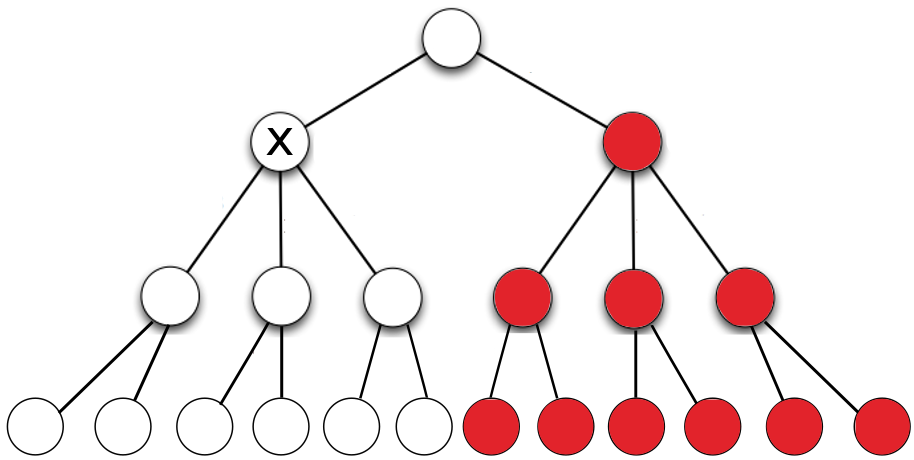}
\caption{Local property. The optimal decision taken by the robot at state $x$ should \emph{not} depend on scenarios that the robot knows cannot occur in the future (shaded in red).
  \label{fig:local property}}
\vspace{-25pt}
\end{wrapfigure}

\noindent {\bf Local property.} At every state of the system, the optimal decision taken by the robot should \emph{not} depend on scenarios that the robot knows cannot occur in the future (see \cite{Shapiro2009, Ruszczynski2010} for a formal discussion\footnote{We note that the local property is sometimes referred to as ``time consistency". However, here we use terminology that is consistent with the dynamic risk measurement literature \cite{Ruszczynski2010}.}). This concept is illustrated in Figure \ref{fig:local property}. The optimal decision taken by the robot at state $x$ should \emph{not} depend on the scenarios shaded in red.

\noindent {\bf Time consistency of risk assessments.} 
Intuitively, time consistency stipulates that if a certain situation is considered less risky than another situation in all states of the world at time-step $k + 1$, then it should also be considered less risky at time-step $k$. Before providing a formal definition, we note that
failure to satisfy time consistency or the local property can lead to ``irrational" behavior, including: (1) intentionally seeking to incur losses \cite{MannorTsitsiklis2011}, or (2) deeming states to be dangerous when in fact they are favorable under any realization of the underlying uncertainty (we discuss an example of this below), or (3) declaring a decision-making problem to be feasible (e.g., satisfying a certain risk threshold) when in fact it is infeasible under \emph{any} possible realization of the uncertainties \cite{RoordaSchumacherEtAl2005}. 

\begin{figure*}
  \begin{center}
    \includegraphics[trim = 0mm 0mm 0mm 0mm, clip, width=0.6\textwidth]{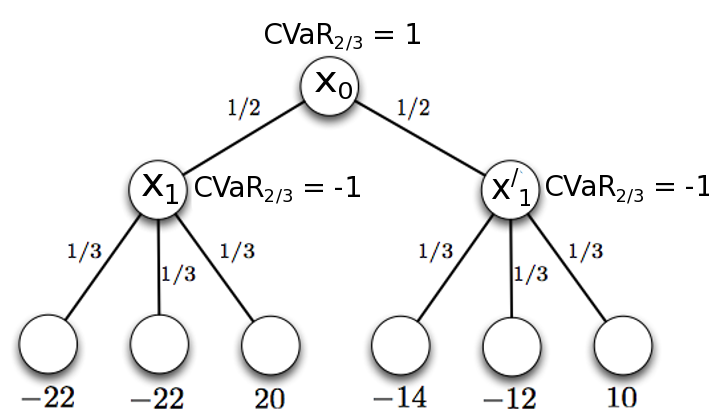}
  \end{center}
  \caption{Failure to satisfy time consistency can lead to ``irrational" behavior. The numbers along the edges represent transition probabilities in the MDP, while the numbers below the terminal nodes represent the terminal costs $c_N(x_N)$. The problem involves a single control policy and there is thus a unique decision tree. The optimal cost appears acceptable at states $x_1$ and $x'_1$, but unacceptable from the perspective of time-step $0$ at state $x_0$. In other words, the decision maker would deem the solution unacceptable even though the solution appears acceptable from the perspective of the second stage ($k=1$) under \emph{any realization of the uncertainties}.
  \label{fig:cvar planning}}
\end{figure*}

As an example, consider the following planning problem with a CVaR cost. Given a Markov Decision Process (MDP) with initial state $x_0$ and time horizon $N \geq 1$, solve:
\begin{flalign} 
	\label{eq:cvar example}
			& & \tau^\star := \min_\pi \hspace*{0.25cm} & \text{CVaR}_\alpha(c_N(x_N)) && \hspace*{-1.75cm}
\end{flalign}
where $\alpha = 2/3$ and $x_N$ is the state at time-step $N$. Consider the scenario tree (based on \cite{ArtznerDelbaenEtAl1999}) in Figure \ref{fig:cvar planning}. Suppose we consider the solution of \eqref{eq:cvar example} acceptable if $\tau^\star \leq 0$. One can then show that the optimization problem (over a single policy) results in an \emph{unacceptable} solution since $\tau^\star$ is positive. However, $\text{CVaR}_\alpha(c_N(x_N)) \leq 0$ is satisfied in every state of the world from the perspective of time-step $k=1$. In other words, the decision maker would deem the solution of \eqref{eq:cvar example} unacceptable even though the solution appears acceptable from the perspective of the second stage ($k=1$) under \emph{any realization of the uncertainties}. 

\begin{figure}
\centering
    \includegraphics[trim = 0mm 0mm 0mm 0mm, clip, width=0.9\textwidth]{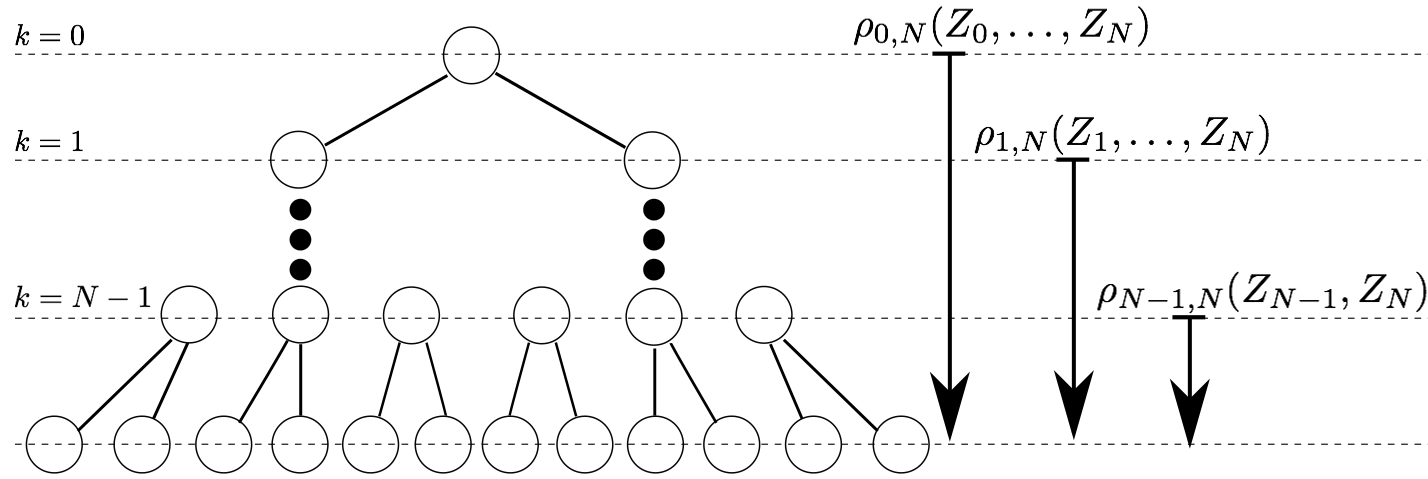}
  \caption{Dynamic risk metrics assess risk at multiple points in time and lead to time-consistent risk assessments. Each $\rho_{k,N}$ maps a future stream of random costs to a risk assessment from the perspective of time-step $k$.
  \label{fig:time consistency}}
\end{figure}

We note that there is nothing special about this example. In general, simply applying a risk metric to the sum of all costs incurred at each time-step does not generally lead to time consistency. In order to obtain time-consistent measures of risk, we need to construct a \emph{sequence} of risk metrics $\{ \rho_{k,N} \}_{k=0}^N$, each mapping a future stream of random costs into a risk assessment at time-step $k$ (visualized in Figure \ref{fig:time consistency}). Such risk metrics are known as \emph{dynamic} risk metrics \cite{Ruszczynski2010} since they assess risk at multiple points in time (in contrast to static risk metrics, which only assess risk from the perspective of the initial stage as in the example above). A dynamic risk metric $\{ \rho_{k,N} \}_{k=0}^N$ is called time-consistent if, for all time-steps $0 \leq l \leq k \leq N$ and all sequences of stage costs $\{ Z_i \}_{i=l}^N$ and $\{ Z'_i \}_{i=l}^N$, the conditions
$$Z_i(\omega) = Z'_i(\omega), \forall \omega \in \Omega, \forall i = l,\dots,k-1, \quad \text{and} \quad \rho_{k,N}(Z_k,\dots,Z_N) \leq \rho_{k,N}(Z'_k,\dots,Z'_N),$$
imply that 
$$\rho_{l,N}(Z_l,\dots,Z_N) \leq \rho_{l,N}(Z'_l,\dots,Z'_N).$$

Remarkably, it can be shown that one can construct a time-consistent risk metric by \emph{compounding} one-step risk metrics \cite{Ruszczynski2010}:
\begin{equation}
\label{eq:composition}
\rho_{k,N} = Z_k + \rho_k(Z_{k+1} + \rho_{k+1}(Z_{k+2} + \dots + \rho_{N-2}(Z_{N-1} + \rho_{N-1}(Z_N))\dots)),
\end{equation}
where the $\rho_k(\cdot)$ functions are a set of single-period risk metrics (satisfying mild technical conditions which are satisfied by distortion risk metrics). These single period metrics assess the risk of a random cost incurred at time-step $k+1$ from the perspective of time-step $k$ (we refer the reader to \cite{Ruszczynski2010} for a more thorough introduction to time consistency of risk metrics). Moreover, under certain mild conditions, \emph{any} time consistent risk metric is of the form \eqref{eq:composition} \cite{Ruszczynski2010}. It can also easily be shown 
that in addition to time-consistency, compounding distortion risk metrics also leads to the local property being satisfied \cite[Chapter 6.8.5]{ShapiroDentchevaEtAl2014}. 

\begin{discussion}[{\bf Time consistency}]
\emph{If one adopts a distortion risk metric as the one-step risk metric that is compounded over time as in Equation \eqref{eq:composition}, one inherits the ``rationality" of the one-step assessments while also ensuring time consistency. This is the form of risk metric we advocate for sequential decision making tasks. However, imposing time consistency still comes at a conceptual cost. In particular, while a risk metric of the form $\rho(\sum_{k=0}^N Z_k)$ is easy to interpret, the composition of one-step metrics in Equation \eqref{eq:composition} stipulated by time consistency is more difficult to interpret. A direction for future work is thus to establish if this tension between time consistency and interpretability is an avoidable one.} 
\end{discussion}

\section{Discussion and Conclusions} 
\label{sec:discussion}

Our goal in this paper has been to provide preliminary directions towards an axiomatic theory of risk for robotics applications. We have advocated properties that risk metrics employed by robots should satisfy in order for them to be considered sensible. These axioms define a class of risk metrics, known as distortion risk metrics, which have been previously used in finance. We further discussed properties that ensure the temporal consistency of risk assessments in sequential decision making tasks.
We end with some questions that highlight areas for future thought in addition to the discussion points highlighted in Discussions 1 and 2 above.

\begin{discussion}[{\bf Further axioms}]
\emph{While we have highlighted a number of axioms that we believe are particularly important, the identification of other axioms is an important direction for future work. These may depend on the particular domain of application. Moreover, for certain applications it may not be necessary to impose all the axioms described here. For example, A4 and A5 (concerning diversification of risks) will generally be relevant to high-level decision making tasks where it is possible to diversify risks and may not be relevant for low-level control tasks where diversification may not be possible. }
\end{discussion}

\begin{discussion}[{\bf Choosing a particular risk metric}]
\emph{For a given application, we may wish to choose a particular risk metric from the class of metrics described here. How should such a metric be chosen? One possibility is to learn a distortion risk metric that explains how humans evaluate risk in the given application domain and then employ the learned risk metric. We describe first steps towards this in \cite{MajumdarSinghEtAl2017}, where we have introduced a framework for \emph{risk-sensitive inverse reinforcement learning} for learning humans' risk preferences from the class of coherent risk metrics.}
\end{discussion}

\begin{discussion}[{\bf Legal frameworks}]
\emph{The question of safety for AI systems has received significant attention recently (see, e.g., \cite{AmodeiOlahEtAl2016} for a recent review). An important component of this discussion has been the consideration of legal frameworks and guidelines that must be placed on AI systems to ensure that they do not pose a threat to our safety. Such considerations for robots such as unmanned aerial vehicles are already extremely pressing for government agencies such as the Federal Aviation Administration (FAA). It is not difficult to imagine a future where the Robot Certification Agency (RCA) is in fact a real entity that certifies the safety of new robotic systems. How can we effectively engage lawmakers and government officials in discussions on how to evaluate risks in robotic applications? 
}
\end{discussion}

Our hope is that the ideas presented in this paper will spur further work on this topic and eventually lead to a convergence upon a particular class of risk metrics that form the standard for assessing risk in robotics. 

\begin{acknowledgement}
The authors were partially supported by the Office of Naval Research, Science of Autonomy Program, under Contract N00014-15-1-2673. 
\end{acknowledgement}



\footnotesize
\bibliographystyle{spmpsci}
\bibliography{../../../bib/main,../../../bib/ASL_papers}

\newcommand{\noopsort}[1]{} \newcommand{\printfirst}[2]{#1}
  \newcommand{\singleletter}[1]{#1} \newcommand{\switchargs}[2]{#2#1}
\begin{thebibliography}{34}
\providecommand{\natexlab}[1]{#1}
\providecommand{\url}[1]{\texttt{#1}}
\expandafter\ifx\csname urlstyle\endcsname\relax
  \providecommand{\doi}[1]{doi: #1}\else
  \providecommand{\doi}{doi: \begingroup \urlstyle{rm}\Url}\fi

\bibitem[Acerbi(2002)]{Acerbi2002}
C.~Acerbi.
\newblock Spectral measures of risk: A coherent representation of subjective
  risk aversion.
\newblock \emph{{Journal of Banking \& Finance}}, 26\penalty0 (7):\penalty0
  1505--1518, 2002.

\bibitem[Allais(1990)]{Allais1990}
M.~Allais.
\newblock Allais paradox.
\newblock In \emph{Utility and Probability}, chapter~2. {Palgrave Macmillan
  UK}, first edition, 1990.

\bibitem[Amodei et~al.(2016)Amodei, Olah, Steinhardt, Christiano, Schulman, and
  Man{\'e}]{AmodeiOlahEtAl2016}
D.~Amodei, C.~Olah, J.~Steinhardt, P.~Christiano, J.~Schulman, and D.~Man{\'e}.
\newblock Concrete problems in ai safety.
\newblock \emph{arXiv preprint arXiv:1606.06565}, 2016.

\bibitem[Artzner et~al.(1999)Artzner, Delbaen, Eber, and
  Heath]{ArtznerDelbaenEtAl1999}
P.~Artzner, F.~Delbaen, J.-M. Eber, and D.~Heath.
\newblock Coherent measures of risk.
\newblock \emph{{Mathematical Finance}}, 9\penalty0 (3):\penalty0 203--228,
  1999.

\bibitem[Bertsimas and Brown(2009)]{BertsimasBrown2009b}
D.~Bertsimas and D.~Brown.
\newblock Constructing uncertainty sets for robust linear optimization.
\newblock \emph{{Operations Research}}, 57\penalty0 (6):\penalty0 1483--1495,
  2009.

\bibitem[Blackmore et~al.(2011)Blackmore, Ono, and
  Williams]{BlackmoreOnoEtAl2011}
L.~Blackmore, M.~Ono, and B.~C. Williams.
\newblock Chance-constrained optimal path planning with obstacles.
\newblock \emph{{IEEE Transactions on Robotics}}, 27\penalty0 (6):\penalty0
  1080--1094, 2011.

\bibitem[Charnes and Cooper(1959)]{CharnesCooper1959}
A.~Charnes and W.~W. Cooper.
\newblock Chance-constrained programming.
\newblock \emph{{Management Science}}, 6\penalty0 (1):\penalty0 73--79, 1959.

\bibitem[Choquet(1954)]{Choquet1954}
G.~Choquet.
\newblock Theory of capacities.
\newblock In \emph{Annales de l'institut Fourier}, 1954.

\bibitem[Delage and Ye(2010)]{DelageYe2010}
E.~Delage and Y.~Ye.
\newblock Distributionally robust optimization under moment uncertainty with
  application to data-driven problems.
\newblock \emph{{Operations Research}}, 58\penalty0 (3):\penalty0 595--612,
  2010.

\bibitem[Du~Toit and Burdick(2011)]{DuToitBurdick2011}
N.~E. Du~Toit and J.~W. Burdick.
\newblock Probabilistic collision checking with chance constraints.
\newblock \emph{{IEEE Transactions on Robotics}}, 27\penalty0 (4):\penalty0
  809--815, 2011.

\bibitem[Ellsberg(1961)]{Ellsberg1961}
D.~Ellsberg.
\newblock Risk, ambiguity, and the savage axioms.
\newblock \emph{{The Quarterly Journal of Economics}}, 75\penalty0
  (4):\penalty0 643--669, 1961.

\bibitem[F{\"o}llmer and Schied(2011)]{FollmerSchied2011}
H.~F{\"o}llmer and A.~Schied.
\newblock \emph{Stochastic Finance: An Introduction in Discrete Time}.
\newblock {Walter de Gruyter}, 2011.

\bibitem[Garc{\'\i}a and Fern{\'a}ndez(2015)]{GarciaFernandez2015}
J.~Garc{\'\i}a and F.~Fern{\'a}ndez.
\newblock A comprehensive survey on safe reinforcement learning.
\newblock \emph{{Journal of Machine Learning Research}}, 16\penalty0
  (1):\penalty0 1437--1480, 2015.

\bibitem[Gilboa and Marinacci(2016)]{GilboaMarinacci2016}
I.~Gilboa and M.~Marinacci.
\newblock Ambiguity and the {Bayesian} paradigm.
\newblock In \emph{Readings in Formal Epistemology}, chapter~21. First edition,
  2016.

\bibitem[Glover and Doyle(1987)]{GloverDoyle1987}
K.~Glover and J.~C. Doyle.
\newblock Relations between {H$_\infty$} and risk sensitive controllers.
\newblock In \emph{Analysis and Optimization of Systems}. {Springer-Verlag},
  1987.

\bibitem[Iancu et~al.(2015)Iancu, Petrik, and Subramanian]{IancuPetrikEtAl2015}
D.~A. Iancu, M.~Petrik, and D.~Subramanian.
\newblock Tight approximations of dynamic risk measures.
\newblock \emph{{Mathematics of Operations Research}}, 40\penalty0
  (3):\penalty0 655--682, 2015.

\bibitem[Kahneman and Tversky(1979)]{KahnemanTversky1979}
D.~Kahneman and A.~Tversky.
\newblock Prospect theory: An analysis of decision under risk.
\newblock \emph{{Econometrica}}, pages 263--291, 1979.

\bibitem[Kuindersma et~al.(2013)Kuindersma, Grupen, and
  Barto]{KuindersmaGrupenEtAl2013}
S.~Kuindersma, R.~Grupen, and A.~Barto.
\newblock Variable risk control via stochastic optimization.
\newblock \emph{{Int.\ Journal of Robotics Research}}, 32\penalty0
  (7):\penalty0 806--825, 2013.

\bibitem[Maccheroni et~al.(2009)Maccheroni, Marinacci, Rustichini, and
  Taboga]{MaccheroniMarinacciEtAl2009}
F.~Maccheroni, M.~Marinacci, A.~Rustichini, and M.~Taboga.
\newblock Portfolio selection with monotone mean-variance preferences.
\newblock \emph{{Mathematical Finance}}, 19\penalty0 (3):\penalty0 487--521,
  2009.

\bibitem[Majumdar et~al.(2017)Majumdar, Singh, Mandlekar, and
  Pavone]{MajumdarSinghEtAl2017}
A.~Majumdar, S.~Singh, A.~Mandlekar, and M.~Pavone.
\newblock Risk-sensitive inverse reinforcement learning via coherent risk
  models.
\newblock In \emph{{Robotics: Science and Systems}}, 2017.
\newblock In press.

\bibitem[Mannor and Tsitsiklis(2011)]{MannorTsitsiklis2011}
S.~Mannor and J.~N. Tsitsiklis.
\newblock Mean-variance optimization in {Markov} decision processes.
\newblock In \emph{{Int.\ Conf.\ on Machine Learning}}, 2011.

\bibitem[on~Banking~Supervision(2014)]{BankingSupervision2014}
B.~C. on~Banking~Supervision.
\newblock \emph{Fundamental Review of the Trading Book: A Revised Market Risk
  Framework}.
\newblock Bank for International Settlements, 2014.

\bibitem[Ono et~al.(2015)Ono, Pavone, Kuwata, and Balaram]{OnoPavoneEtAl2015}
M.~Ono, M.~Pavone, Y.~Kuwata, and J.~Balaram.
\newblock Chance-constrained dynamic programming with application to risk-aware
  robotic space exploration.
\newblock \emph{{Autonomous Robots}}, 39\penalty0 (4):\penalty0 555--571, 2015.

\bibitem[Rockafellar and Uryasev(2000)]{RockafellarUryasev2000}
R.~T. Rockafellar and S.~Uryasev.
\newblock Optimization of conditional value-at-risk.
\newblock \emph{{Journal of Risk}}, 2:\penalty0 21--41, 2000.

\bibitem[Roorda et~al.(2005)Roorda, Schumacher, and
  Engwerda]{RoordaSchumacherEtAl2005}
B.~Roorda, J.~M. Schumacher, and J.~Engwerda.
\newblock Coherent acceptability measures in multi-period models.
\newblock \emph{{Mathematical Finance}}, 15\penalty0 (4):\penalty0 589--612,
  2005.

\bibitem[{Ruszczy\'nski}(2010)]{Ruszczynski2010}
A.~{Ruszczy\'nski}.
\newblock Risk-averse dynamic programming for {Markov} decision process.
\newblock \emph{{Mathematical Programming}}, 125\penalty0 (2):\penalty0
  235--261, 2010.

\bibitem[Schmeidler(1986)]{Schmeidler1986}
D.~Schmeidler.
\newblock Integral representation without additivity.
\newblock \emph{{Proceedings of the American Mathematical Society}},
  97\penalty0 (2):\penalty0 255--261, 1986.

\bibitem[Shapiro(2009)]{Shapiro2009}
A.~Shapiro.
\newblock On a time consistency concept in risk averse multi-stage stochastic
  programming.
\newblock \emph{{Operations Research Letters}}, 37\penalty0 (3):\penalty0
  143--147, 2009.

\bibitem[Shapiro et~al.(2014)Shapiro, Dentcheva, and
  Ruszczy{\'n}ski]{ShapiroDentchevaEtAl2014}
A.~Shapiro, D.~Dentcheva, and A.~Ruszczy{\'n}ski.
\newblock \emph{Lectures on stochastic programming: Modeling and theory}.
\newblock {SIAM}, second edition, 2014.

\bibitem[Summers et~al.(2015)Summers, Warrington, Morari, and
  Lygeros]{SummersWarringtonEtAl2015}
T.~Summers, J.~Warrington, M.~Morari, and J.~Lygeros.
\newblock Stochastic optimal power flow based on conditional value at risk and
  distributional robustness.
\newblock \emph{International Journal of Electrical Power \& Energy Systems},
  72:\penalty0 116 -- 125, 2015.

\bibitem[von Neumann and Morgenstern(1944)]{NeumannMorgenstern1944}
J.~von Neumann and O.~Morgenstern.
\newblock \emph{Theory of Games and Economic Behavior}.
\newblock {Princeton University Press}, 1944.

\bibitem[Wang(2000)]{Wang2000}
S.~Wang.
\newblock A class of distortion operators for pricing financial and insurance
  risks.
\newblock \emph{Journal of Risk and Insurance}, 67\penalty0 (1):\penalty0
  15--36, 2000.

\bibitem[Whittle(1981)]{Whittle1981}
P.~Whittle.
\newblock Risk-sensitive linear/quadratic/gaussian control.
\newblock \emph{{Advances in Applied Probability}}, 13\penalty0 (4):\penalty0
  764--777, 1981.

\bibitem[Xu and Mannor(2010)]{XuMannor2010}
H.~Xu and S.~Mannor.
\newblock Distributionally robust {Markov} decision processes.
\newblock In \emph{{Advances in Neural Information Processing Systems}}, 2010.

\end{thebibliography}

\end{document}